\documentclass[runningheads]{llncs}

\usepackage{eccv}

\usepackage{eccvabbrv}

\usepackage{graphicx}
\usepackage{booktabs}

\usepackage[accsupp]{axessibility}  % Improves PDF readability for those with disabilities.
\usepackage{mathrsfs}
\usepackage{bbding}
\usepackage{color}
\usepackage{booktabs}
\usepackage{listings}
 
\newcommand{\Epsilon}{\mathcal{E}} % 定义大写Epsilon命令
\usepackage{multicol}
\usepackage{multirow}
\usepackage{makecell}

\usepackage[pagebackref,breaklinks,colorlinks,citecolor=eccvblue]{hyperref}
\usepackage{orcidlink}

\begin{document}

% ---------------------------------------------------------------
% TODO REVIEW: Replace with your title
\title{DiffuMatting: Synthesizing Arbitrary Objects with Matting-level Annotation} 

% TODO REVIEW: If the paper title is too long for the running head, you can set
% an abbreviated paper title here. If not, comment out.
\titlerunning{DiffuMatting}

% TODO FINAL: Replace with your author list. 
% Include the authors' OCRID for the camera-ready version, if at all possible.
%%\dag
\author{Xiaobin Hu\inst{1}{\thanks{$\star$ these authors contributed equally to this work. \\
$\dag$ indicates equally corresponding authors. }} \and
Xu Peng\inst{1,2\footnotemark[1]}  \and Donghao Luo\inst{1 \dag}
% \thanks{equally corresponding authors}
\and Xiaozhong Ji\inst{1} \and Jinlong Peng\inst{1} \and Zhengkai Jiang\inst{1} \and Jiangning Zhang\inst{1} \and \\
Taisong Jin\inst{2 \dag} \and Chengjie Wang\inst{1} \and Rongrong Ji\inst{2}}

% Third Author\inst{3}\orcidlink{2222--3333-4444-5555}
% TODO FINAL: Replace with an abbreviated list of authors.
\authorrunning{X.~Hu et al.}
% First names are abbreviated in the running head.
% If there are more than two authors, 'et al.' is used.

% TODO FINAL: Replace with your institution list.
\institute{Tencent Youtu Lab \and
Key Laboratory of Multimedia Trusted Perception and Efficient Computing. \\Ministry of Education of China, Xiamen University. 361005, PR. China. \\
% \email{lncs@springer.com}\\
% \url{http://www.springer.com/gp/computer-science/lncs} \and
% ABC Institute, Rupert-Karls-University Heidelberg, Heidelberg, Germany\\
% \email{\{abc,lncs\}@uni-heidelberg.de}
}

\maketitle

\begin{abstract}
Due to the difficulty and labor-consuming nature of getting highly accurate or matting annotations, there only exists a limited amount of highly accurate labels available to the public.  
To tackle this challenge, we propose a DiffuMatting which inherits the strong Everything generation ability of diffusion and endows the power of `matting anything'. Our DiffuMatting can \textit{1).} act as an anything matting factory with high accurate annotations
\textit{2).} be well-compatible with community LoRAs or various conditional control approaches to achieve the community-friendly art design and controllable generation. 
Specifically, inspired by green-screen-matting, we aim to teach the diffusion model to paint on a fixed green screen canvas. To this end, a large-scale green-screen dataset (Green100K) is collected as a training dataset for DiffuMatting. 
Secondly, a green background control loss is proposed to keep the drawing board as a pure green color to distinguish the foreground and background. 
To ensure the synthesized object has more edge details, a detailed-enhancement of transition boundary loss is proposed as a guideline to generate objects with more complicated edge structures. 
Aiming to simultaneously generate the object and its matting annotation, 
we build a matting head to make a green-color removal in the latent space of the VAE decoder.
Our DiffuMatting shows several potential applications (\textit{e.g.,} matting-data generator, community-friendly art design and controllable generation). As a matting-data generator, DiffuMatting synthesizes general object 
and portrait matting sets, 
effectively reducing the relative MSE error by 15.4\% in General Object Matting. 
% and 11.4\% in Portrait Matting tasks. 
The dataset is released in our project page at \url{https://diffumatting.github.io}.

% For downstream general object matting and portrait matting tasks, we synthes 
% 10K human portrait is synthesized by our DiffuMatting to verify its importance for downstream portrait matting tasks. 
% From quantitative results, the existing matting methods trained on synthetic data of DiffuMatting and real set lower MSE error by 14.3\% than the results of only the real set.
% \vspace{-2mm}
\end{abstract}
% \vspace{-5mm}

\begin{figure}[t]
\centering
\includegraphics[width=1\textwidth]{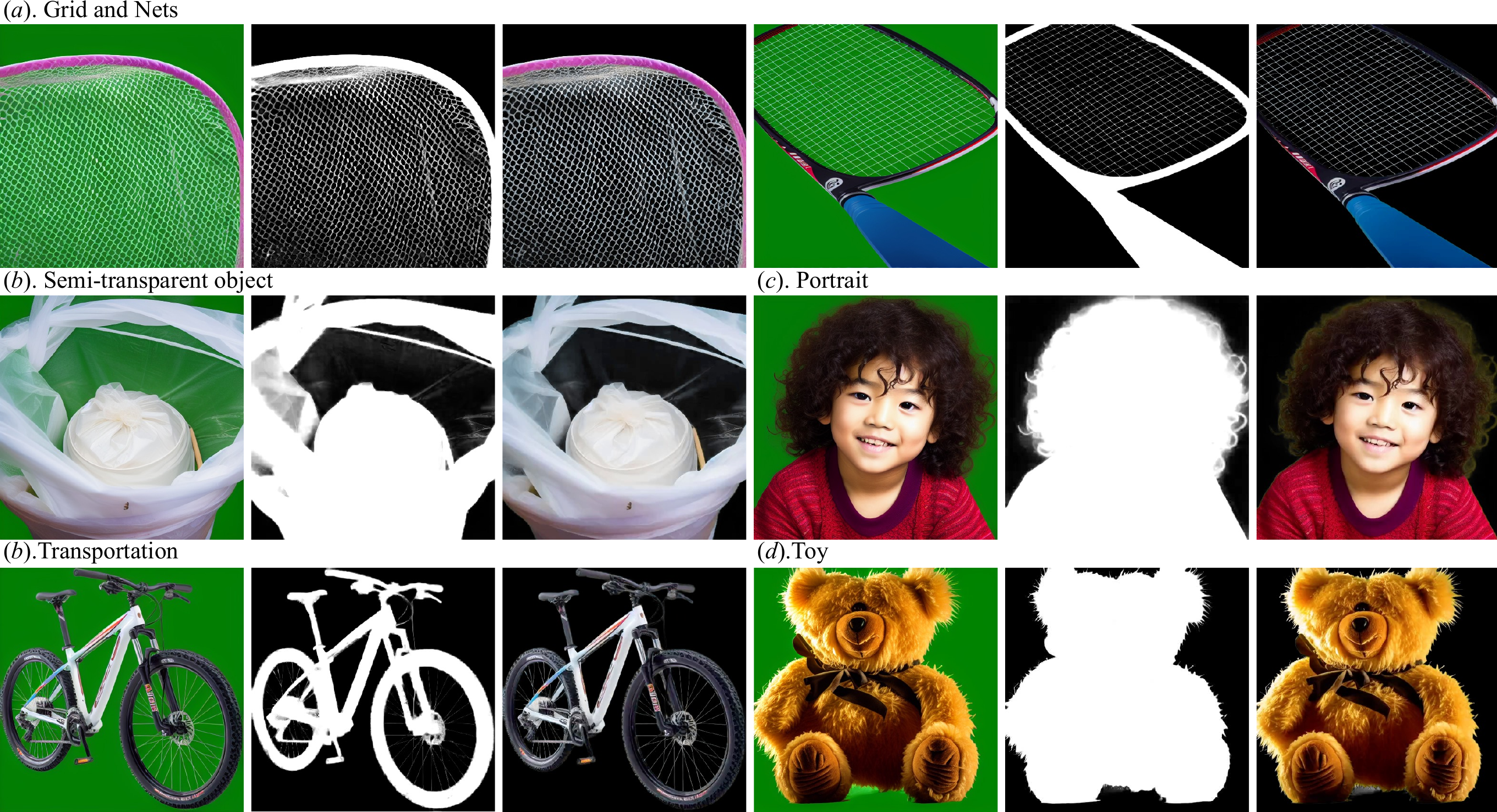}
\vspace{-7mm}
\caption{Green-screen objects with matting-level annotations generation by DiffuMatting, including nets, grid and semitransparent tough objects and extended to almost any class (\textit{e.g.,} Transportation, Architecture, Toy).
% without any parameters fine-tuning.
}
\label{fig:abs}
\vspace{-3mm}
\end{figure}

\section{Introduction}
\vspace{-2mm}
\label{sec:intro}
High-accurate annotation (including segmentation of complex topological structures or matting objects) aims to provide highly delicate labels for objects. 
With the development of network architecture, the performance of accurate annotation has been boosted to some extent, and it provides more geometrically descriptions or transparency of the objects needed in many applications, augmented reality (AR) \cite{qin2021boundary}, image editing \cite{goferman2011context}, 3D reconstruction \cite{liu2021fully}, image composition \cite{zhu2021barbershop}. 
Besides, considering the laborious labeling process and unacceptable cost, there only exists limited training pairs, \textit{e.g.,} Adobe matting dataset \cite{xu2017deep} with only 493 foreground matting objects and Dinstinctions-646 \cite{qiao2020attention} composed of 646 foreground images with manually annotated alpha mattes. It remains an open question of how to construct a highly accurate and efficient data-factory to simultaneously synthesize the delicate objects and their matting-level annotations, which also facilitating the composition and creation of content.
%%
% their exceptional performance
% in synthesizing high-quality and diverse images

Recently, the emergence of diffusion models has shown its milestones in the exceptional generative ability to synthesize high-quality and diverse images. 
It unavoidably raises the consideration if the generative capability benefits the downstream tasks, especially on matting, segmentation, and image composition. 
% There exists some work \cite{he2022synthetic,sauer2022stylegan,bansal2023leaving} on exploring the synthetic images to enhance the semantic understanding which effectively improves the classification accuracy even without the accurate annotations. 
There exists some work \cite{he2022synthetic,sauer2022stylegan,bansal2023leaving} on exploring the synthetic images to effectively improves the classification accuracy. 
As a more challenging task, segmentation requires both the synthetic images and its pixel-level aligned annotations. 
The segmentation data-generator can be categorized as `conditional' and `unconditional' manners. 
Conditional generation \cite{yang2023freemask,wang2021image} requires semantic masks as the conditional inputs, and the generation scope is restricted within the semantic layouts.
It fails to generate arbitrary objects (`Anything') but is only limited in the close domain with the training set. 
In contrast, unconditional generations \cite{wu2023diffumask,shenoda2023diffugen} can largely extend the generation space not limited to the same domain of semantic layouts. 
% Previous class-prior training methods \cite{baranchuk2021label,park2023learning} mainly learn the mapping between the image and mask in the specified class, and damages the anything generation ability of stable diffusion. 
% Previous class-prior training methods \cite{baranchuk2021label,park2023learning} mainly learn the mapping between the image and mask in the specified class, and damages the anything generation ability of stable diffusion. 
But it is still facing several challenges: 
\textit{1).} It still does not fully exploit the Anything Generation of denoising diffusion probabilistic models, not beyond the class-priors in the training setting.  
% \textit{2).} It needs manual annotations \cite{zhang2021datasetgan,li2022bigdatasetgan} or suffers from the pixel-level misalignment between the synthetic images and annotations \cite{ding2023large}.
% \textit{2).} It suffers from the pixel-level misalignment between the synthetic images and annotations \cite{ding2023large}.
\textit{2).} It fails to generate the matting-level annotations.

% where the delicate object is usually labeled 30 minutes/per image \cite{qin2022highly}.

\begin{figure}[t!]
\centering
\includegraphics[width=1\linewidth]{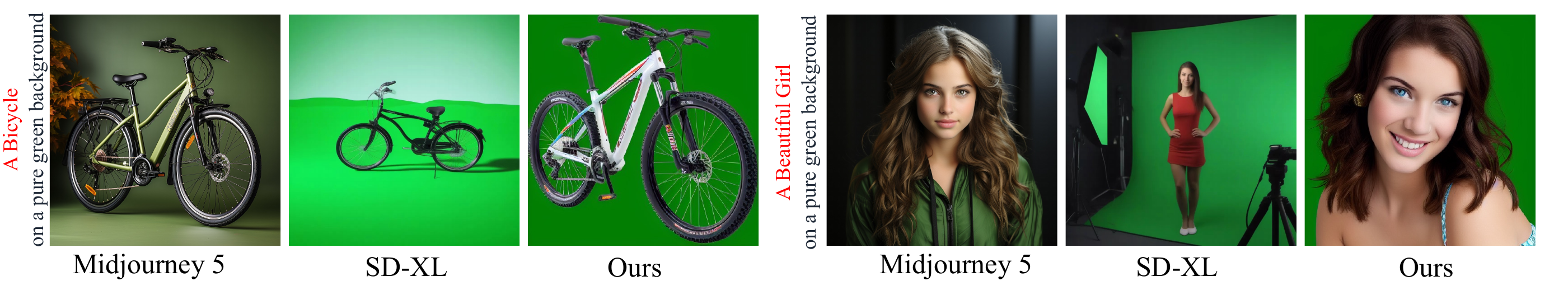}
\vspace{-6mm}
\caption{
Visual performance of our DiffuMatting on green-screen object generation in comparison with SOTA Midjourney and SD-XL models, and these models have difficulties in consistently generating objects on the pure green-screen.
% Midjourney and SD-XL fails to steadily generate arbitrary ultra-detailed object on the green-screen. 
}
\vspace{-5mm}
\label{fig:midsdxl_green}
\end{figure}

\textit{To tackle the first challenge of Anything Generation}, inspired by the green-screen matting, we ingeniously teach the diffusion model to paint on a fixed pure green screen canvas, which easier to distinguish the foreground and matting-level annotations. 
In other words, such a simple way embeds the object and high-accurate mask into the same green-screen image. 
This setting not only inherits the generic knowledge of the large-scale diffusion model by avoiding converging to class-prior training domain but also endows the diffusion model to naturally learn the background (green canvas) and foreground objects. 
% we adopt powerful pretrained stable diffusion models which acquire the generic knowledge to excavate the exceptional Anything Generation ability. 
% The text-based prompt is the easiest control gate to generate the anything object without requiring other conditions. 
% Previous image-to-mask paired training damages the generation ability of stable diffusion and converges it to the image-to-mask domain. 
% Inspired by the green-screen matting, we ingeniously teach the diffusion model to paint on a fixed pure green screen canvas, which easier to distinguish the foreground and matting-level annotations. 
% In other words, such a simple way embeds the object and high-accurate mask into the same green-screen image. 
% This setting not only inherits the generic knowledge of the large-scale diffusion model by avoiding converging to class-prior training domain but also endows the diffusion model to naturally learn the background (green canvas) and foreground objects.
% However, even with such a setting, there are still several challenges and cor:
% Considering that the mainstream of SOTA diffusion models (\textit{e.g.,} DALLE3, SDXL \cite{podell2023sdxl}, Midjourney) fails to steadily generate arbitrary ultra-detailed objects on the green canvas shown in Fig. \ref{fig:midsdxl_green}, this means that no existing stable and robust green-canvas-based diffusion model is available.
Given that current SOTA diffusion models, (\textit{e.g.,} DALLE3, SDXL \cite{podell2023sdxl}, Midjourney), struggle to consistently generate arbitrary ultra-detailed objects on the green canvas as depicted in Fig. \ref{fig:midsdxl_green}, it underscores the absence of a stable, robust diffusion model tailored for green-canvas-based scenarios. Thus, a generic green-canvas-based dataset Green100k is collected including 100,000 high-resolution images paired with high-accurate annotations, which makes it possible to train our DiffuMatting diffusion models. Additionally, a control loss with a green background is proposed to leverage cross-attention features associated with the `green' color, ensuring the preservation of a pure and stable canvas background. 
% For the second challenge of high-accurate matting annotations, a detailed-enhancement of transition boundary loss is proposed to generate more high-detailed objects the same as in matting task and also avoid transition boundary crash. 
% The mask is only generated based on the latent space to synthesize paired well-aligned image-mask without sacrificing the Anything generation ability, which avoids being involved in the coarse-level noise-based synthesis processes. In such a setting, the sub-pixel well-aligned annotation synthesis can be simplified as the matting-header operation in the decoder and the matting-level post-processing.
\textit{To address the second challenge of achieving high-accuracy matting annotations}, we propose a detailed-enhancement of transition boundary loss. This approach aims to generate objects with higher detail, akin to those in matting tasks, while also preventing transition boundary crashes. The mask generation is solely based on the latent space, allowing for the synthesis of well-aligned image-mask pairs without compromising the ability to generate anything. This approach circumvents the need for involvement in coarse-level, noise-based synthesis processes. In such a setting, the synthesis of sub-pixel well-aligned annotations can be simplified as a matting-header operation in the decoder, followed by matting-level post-processing.
Our contributions are summarized as:
\vspace{-1mm}
\begin{itemize}
    % \item \textcolor{red}{revise} A generic green-canvas-based dataset Green100k is collected including 100,000 high-resolution images paired with matting annotation, which makes it possible to train our DiffuMatting diffusion models. \textcolor{red}{remove this line and not as the motivation}
    % two loss
    \item 
    % We propose a green-background control loss  to utilize the cross-attention features tied with the `green' to keep the canvas-background pure and stable. Additionally, 
    % a detailed-enhancement of transition boundary loss is proposed to enrich the boundary and avoid transition boundary crash.
    % We propose two novel loss functions:  a green-background control loss and the transition boundary loss. The former leverage cross-attention features associated with 'green' to maintain a pure and stable canvas background, while the latter aimed at enriching boundary detail and preventing transition boundary crashes.
    
    % We introduce two novel loss functions: a green-background control loss, leveraging cross-attention features associated with 'green' to maintain a pure and stable canvas background, and a detailed-enhancement of transition boundary loss, designed to enhance boundary detail while preventing transition boundary crashes.
    We propose two novel loss functions: the green-background control loss and the transition boundary loss. The former leverages `green' cross-attention features to ensure a stable canvas background, while the latter focuses on enhancing boundary detail and avoiding transition boundary crashes.
    %
    % \item To well align the image-mask pairs and keep the anything generation ability, a matting head is built to make a green-color removal on the latent space of VAE decoder which avoids being involved in the coarse-level noise-based synthesis processes. 
    %% to arrive matting-accuracy 
    \item To produce annotations at the matting level, Green100K is collected for training and a matting head is built to make a green-color removal on the latent space of VAE decoder which avoids being involved in the coarse-level noise-based synthesis processes. 
    \item As a role of matting-data factor, we synthesize the general object and portrait matting set respectively and demonstrate that the inclusion of synthesized data leads to a reduction of 15.4\% relative MSE error in General Object Matting.
    % and 11.4\% relative MSE error in Portrait Matting task.
    
    \item DiffuMatting is well-compatible with various community LoRAs and existing control models (\textit{e.g.,} ControlNet) without additional training or allows users to customize specific styles images with matting annotations.

    % DiffuMatting has a strong anything-generation ability shown in Fig. \ref{fig:abs}, which 
    % % well-generate objects 
    % % largely beyond the training set and 
    % covers most of the objects in real and virtual life. 
    % The application shows our DiffuMatting model can generate arbitrary target object and matting annotation in the scene reconstruction of image composition. 
    % As a role of matting-data factor, we synthesize the general object and portrait matting set respectively and demonstrate that the inclusion of synthesized data leads to a reduction of 15.4\% relative MSE error in General Object Matting and 11.4\% relative MSE error in Portrait Matting task.

    %
%     averagely lowers the relative MAE error by
% 16.9% and improves relative Fw
% β by 7.5% on four datasets
    % 3d modeling vr composition 
\end{itemize}

% \vspace{-2mm}
\section{Related Work}
% \vspace{-1mm}
\label{sec:formatting}

%-------------------------------------------------------------------------

%%%
%%%1.Text-to-image diffusion models
%%%2.Synthetic Dataset Generation.
%%%3.Matting and high-accurare segmentation.

\vspace{-1mm}
\subsection{Text-to-image Diffusion Models}
% Diffusion models[] have recently achieved remarkable success in image generation, driving advancements in various applications and fields. Their powerful performance has significantly propelled the field of text-guided image synthesis[] forward. In particular, large-scale text-to-image diffusion models, trained on extensive text-image pair datasets, have set new benchmarks. Notable examples include Stable diffusion[], Imagen[] and DALL-E3[]. 
%%%
Diffusion models have recently demonstrated remarkable success in the field of image generation, leading to advancements in various applications and disciplines. 
Their impressive performance has greatly boosted the progress of text-guided image synthesis. 
Concretely, the development of large-scale text-to-image diffusion models, trained on extensive datasets of text-image pairs, has established new benchmarks in this domain. 
Prominent examples of such models include Stable diffusion \cite{rombach2022high}, Imagen \cite{saharia2022photorealistic}, DALL-E3, and other variants \cite{li2023bbdm,gu2022vector,peebles2023scalable,ho2022classifier,karras2022elucidating}.
However, these SOTA diffusion models fail to steadily generate arbitrary ultra-detailed objects on the green canvas via a text-based prompt control. 
Our work inherits the generic knowledge of large-scale models by avoiding the domain restriction of image-mask training pairs. 
Additionally, by introducing a fixed green screen canvas, it enables the diffusion model to inherently learn both the background (green canvas) and foreground. 
% objects.

% By avoiding convergence to the domain of training image-mask pairs, this setting inherits the generic knowledge of large-scale diffusion models. Additionally, it enables the diffusion model to naturally learn both the background (green canvas) and foreground objects
%%
% the mainstream of SOTA diffusion models (e.g., DALLE3, SDXL, Midjourney) fails to  steadily generate any ultra-detailed object  on the green canvas. It means that no existing green-canvas-based diffusion model is available.
%
% This setting not only inherits the generic knowledge of large-scale diffusion model by avoiding converging to training image-mask-pair domain, but also endows the diffusion model to naturally learn the background (green canvas) and foreground objects.
%-------------------------------------------------------------------------

\vspace{-2mm}
\subsection{Synthetic Data Generation}
\vspace{-1mm}
Previous studies \cite{kar2019meta,devaranjan2021unsupervised} on dataset synthesis primarily employ 3D scene graphs (graphics engines) to generate images along with their corresponding labels. Nevertheless, these datasets often demonstrate a domain gap compared to real-world datasets, encompassing discrepancies in both appearance and content. A generative adversarial network leverages image-to-image translation to mind this gap in appearance and content. To generate infinite synthetic images and mask correspondingly, DatasetGAN \cite{zhang2021datasetgan} employed a limited number of labeled real images to train a segmentation mask decoder. Building upon DatasetGAN, BigDatasetGAN \cite{li2022bigdatasetgan} expanded the class diversity to the scale of ImageNet, generating 1k classes with manually annotated 5 images per class. 
With the emergence of diffusion models, there have been some initial attempts to apply them to generate synthesis images for downstream tasks, 
\textit{e.g.,} classification task \cite{he2022synthetic,azizi2023synthetic}, synthetic ImageNet \cite{sariyildiz2023fake}, altering the color \cite{trabucco2023effective}, segmentation \cite{wang2022semantic,baranchuk2021label,park2023learning,wu2023diffumask,shenoda2023diffugen}.
FreeMask \cite{yang2023freemask} synthesizes training images on the scene understanding with the conditions of semantic mask provided by realistic dataset. 
Also in these works \cite{baranchuk2021label,yang2023freemask,azadi2019semantic,le2021semantic}, they also require the condition of the semantic layout, leading to tricky generate user-specified subjects. Ideally, the text-conditioned generation model is capable of relaxing the strict constraints, and synthesizing arbitrary objects via text phrase, which makes it potential to generate `Anything'. DiffuMask \cite{wu2023diffumask} exploits the potential of the cross-attention map guided by text phrases to synthesize the predefined classes in the training set. Different from the above methods, our DiffuMatting can generate matting-level sub-pixel annotations which are much more expensive and labor-consuming than pixel-level masks. In addition, our DiffuMatting is not restricted by the training set but be largely extended to generate the most objects in real and virtual life.

\begin{figure*}[t!]
\centering
\includegraphics[width=1\textwidth]{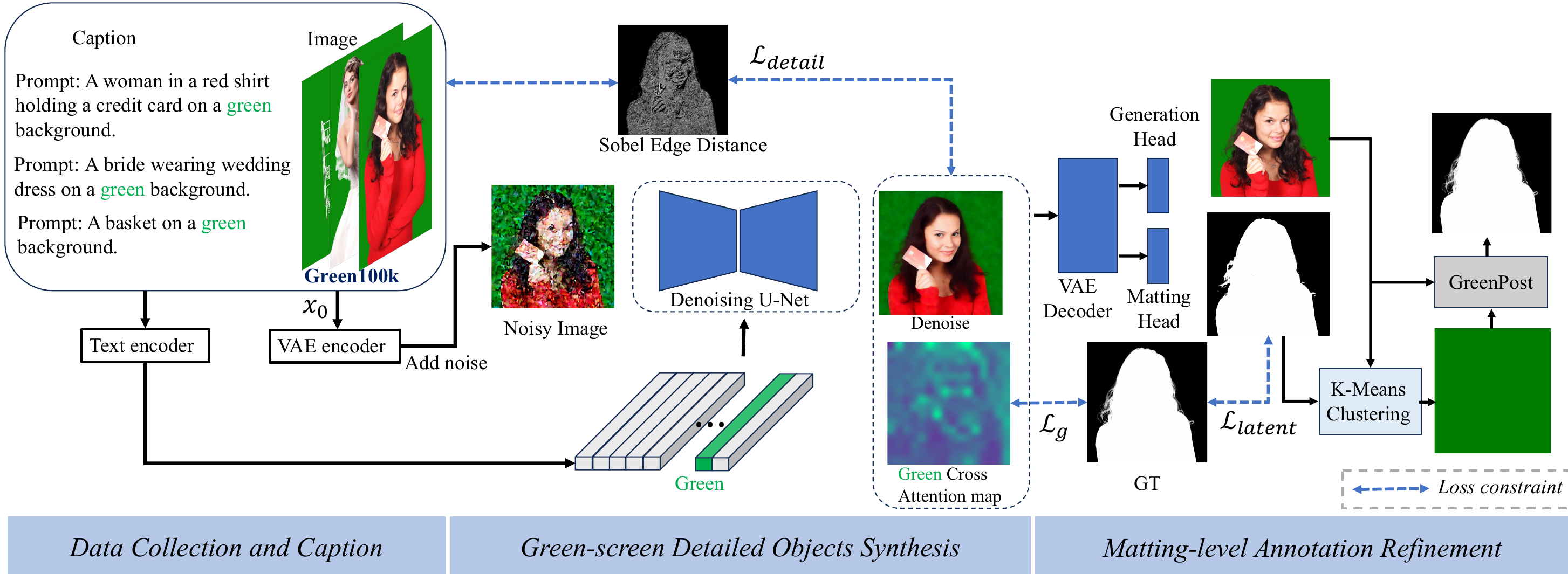}
\vspace{-3mm}
\caption{An overview of Our DiffuMatting Network. Our DiffuMatting mainly consists of Green100k data collection and caption, green-screen detailed objects synthesis assisted by the green-background control loss $\mathcal{L}_{g}$ and the detailed-enhancement loss of transition boundary  $\mathcal{L}_{detail}$, and matting-level annotation refinement via a matting-head in VAE latent space constrained by  $\mathcal{L}_{latent}$ and 
% $\mathbf{GreenPost}$.
{GreenPost}.
}\label{fig:fig3_diffumat}
% \vspace{-6mm}
\end{figure*}

\vspace{-3mm}
\subsection{Matting-level Dataset}
Image matting refers to the precise estimation of the foreground object in images and videos. This technique holds significant importance in image and video editing applications, especially in the realm of film production where it is utilized to create captivating visual effects.
However, considering the significant labor on matting-level annotations, Composition-1k \cite{xu2017deep}, the most widely used dataset, only contains 431 foregrounds for training and 20 foregrounds for testing. 
To improve the versatility and robustness of the matting, Distinctions-646 dataset \cite{qiao2020attention} is proposed with only 646 distinct FG images, AIM-500 \cite{li2021deep} comprising 500 high-quality real natural images, P3M-500 \cite{li2021privacy} with 500 high-fidelity portraits and AM-2k \cite{li2022bridging} with 2,000 animal images. 
In summary, the amount of matting-level annotation is extremely rare compared with other pixel-level segmentations 11 million well-annotated images in SAM \cite{kirillov2023segment}. 
It is urgent to investigate a much cheaper and time-saving manner to synthesize abundant and almost unlimited images and matting-level annotations for any object.

% \vspace{-3mm}
\section{Methodology}
% In this paper, we investigate simultaneously to generate the images and the matting-level annotations with the descriptions of text phrases, breaking the constraint of the training domain. The overview structure of the DiffuMatting is shown in Fig. \ref{fig:fig3_diffumat}.
% \vspace{-2mm}
\noindent\textbf{Motivation and Applications.} 
Despite the widespread use of large-scale models for image generation, there is little research focus on matting-level transparent image generation.  
Incorporating transparency is extensively used in the majority of visual content editing software and workflows to facilitate the composition and creation of content, which is significantly needed by the commercial market. In addition, the scarcity of matting-level data hinders the development of matting algorithms towards high accuracy and strong robustness. 
Driven by these demands, we propose DiffuMatting to generate the 4-channels images with the high-accurate matting channels for the following core-meaning applications: 
\textit{1)}. Data factory. 
% Considering the difficulty and labor-consuming nature of getting highly accurate or matting annotations, 
DiffuMatting can act as a data factory for downstream matting tasks in Fig. \ref{fig:general_matting}.
% and \ref{fig:portrait_matting}.
% \textit{1)}. Image composition. Our DiffuMatting can generate, copy, and paste the 4-channels with the ultra-detailed objects and annotations into the desired scene in Fig. \ref{fig:composition}.
\textit{2)}. Community-friendly art design and controllable generation. DiffuMatting exhibits excellent compatibility with community LoRAs and various conditional control approaches in Fig. \ref{fig:art_design_lora_exp} and Fig. \ref{fig:art_design_control}. It also supports the image composition in desired scenario in Fig. \ref{fig:composition}.
% image composition in Fig. \ref{fig:composition}.

% \textit{3)}. User-specified style customization. DiffuMatting exhibits excellent compatibility with LoRA models, offering users the potential to customize RGBA 4-channel images with specific styles in Fig. \ref{fig:art_design}.
% \textit{4)}. Community-friendly art design. DiffuMatting can well integrate into various community LoRA models without requiring any additional training process in Fig. \ref{fig:art_design_lora_exp}.
% \textit{5)}. Controllable image generation. The current control models (\textit{e.g.,} ControlNet) can be directly utilized in conjunction with our DiffuMatting to enrich the functionality of controllable image editing in Fig. \ref{fig:art_design_control}.

% \textit{4)}. Materials for 3D modeling. 3D models built upon the masks sampled from DiffuMatting by the “Extrude” operation in Blender. 

%%整体归纳一下结构
\vspace{-3mm}
\subsection{Data Collection of Green100k Dataset}\label{section:DataCollectionparts}
\vspace{-1mm}
To address the dataset issue, we build a matting-level Green100k Dataset with 10,0000 green-screen-background objects and its high-accurate annotations which are mainly collected from public matting-level datasets and self-shoot green-screen portrait videos.
Specifically, for self-collected portrait videos, we totally collected 20 portrait videos on the green-screen background.  We annotate the above green-screen images in a matting manner with Adobe Premiere Pro Chromakey and Adobe Photoshop.
To avoid some consecutive video frames, we sample one frame every 20 consecutive frames and the amount of self-collected images arrives at 1,1963.
Considering that DiffuMatting is a general object generation model inheriting everything generation ability, we aim to increase the diversity of the collected data set and then include a small part of the high-accurate salient object dataset. 
% we collect a small part of high-accurate salient object dataset 
% To increase the diversity of dataset, 
Specifically, we collect the public available matting-level or high-accurate 
dataset including Adobe matting \cite{xu2017deep} (454 images), Distinction646 \cite{qiao2020attention} (590 images), P3M \cite{li2021privacy} (481 images), PPM \cite{ke2022modnet} (98 images), Video240k matting-set \cite{lin2021real} (5,0000 images), AM2K \cite{li2022bridging} (1942 images), DIS5K \cite{qin2022highly} (5314 images), HRSOD \cite{zeng2019towards} (1918 images), MSRA10k \cite{hou2017deeply} (9884 images), DUTS \cite{wang2017learning} (10426 images) and Thinobject5k \cite{liew2021deep} (5594 images) dataset. The objects of the public matting-level dataset are synthesized into a green-screen background based on the corresponding accurate annotations.
%%如何打标签+如何标注
For the training of the stable diffusion model, we use the instruct-blip algorithm \cite{instructblip} to caption our Green100k dataset. To avoid the misalignment between the content and captions, a great labor of manual correction is implemented on all data.
%%
% \begin{figure}[t!]
% \centering
% \includegraphics[width=0.95\linewidth]{figures/dataset.png}
% \vspace{-3mm}
% \caption{
% Groups of our Green100k Dataset. Due to the pixel-level annotation of Video240K, only a part of Video240K is introduced in our Green100k.
% } 
% \label{fig:piechart}
% \vspace{-3mm}
% \end{figure}
\vspace{-3mm}
\subsection{Preliminaries of Cross-attention Mechanism}
\vspace{-1mm}
Stable diffusion is a probabilistic model specifically designed to estimate a data distribution p(x) by iteratively reducing noise in a normally distributed variable. Given an input image $x_{0}$, the noise estimation process is defined as:
\vspace{-2mm}
\begin{equation}
\vspace{-1mm}
\mathcal{L}_{noise} = \mathbb{E}_{z \sim \Epsilon(x), C, \epsilon \sim \mathcal{N}(0,1), t} \left[||\epsilon - \epsilon_{\theta}(z_t,t,C)||^2_{2} \right],
\vspace{-1mm}
\end{equation}
where a Variational AutoEncoder $\Epsilon$ compress  input image $x$ into low-dimension latent space $z$. A conditional U-Net denoiser $\epsilon_{\theta}$ aims to estimate noise $\theta$ in  latent space with the aid of timestep $t$, $t$-th noisy latent representation $z_{t}$, and other text-prompt conditions $C$ extracted by text-encoder. 

The cross-attention mechanism is implemented to fuse the visual and textual embedding via:
\vspace{-2mm}
\begin{equation}
\vspace{-1mm}
\mathcal{A}=\mathrm{Softmax}\left(\frac{{\ell}_{Q}(\varphi(z_{t}))\ell_{K}(\tau_{\theta}(\mathcal{P}))^{T}}{\sqrt{d}}\right),
\vspace{-1mm}
\end{equation}
where $\ell_{Q}(\varphi(z_{t})$ is the operation to flatten and linearly project deep spatial features of noisy image $\varphi(z_{t})$ to Query vector, and $\ell_{K}(\tau_{\theta}(\mathcal{P}))$  is to lineraly project prompt $\mathcal{P}$ into textural embedding $\tau_{\theta}(\mathcal{P})$ as a Key matrix ($K$). $d$ is latent projection dimension, $\ell_{K}$ and $\ell_{Q}$ are linearly learnable projection matrices.
% The visual features of the noise images $\varphi(z_{t}) \in \mathbb{R}^{H \times W \times C}$ are flatted and lin

% $\Epsilon$  
% \begin{equation}
%     \hat{z}_0 = \dfrac{z_t-\sqrt{1-\bar{\alpha}_t}\epsilon_{\theta}}{\sqrt{\bar{\alpha}_t}}, t< \delta T,
% \end{equation}
% Stable diffusion mainly consist of three componements 
% Diffusion Models [82] are probabilistic models 
\vspace{-3mm}
\subsection{Green-background Control}
\vspace{-1mm}
The overview structure of DiffuMatting is shown in Fig. \ref{fig:fig3_diffumat}. Given Green100k dataset, we need strong priors to restrict fore- and background to ensure the clean green screen of background zone. 
Intuitively, we can directly tie the main object of prompts with cross-attention mask to split fore- and background. 
However, to empower model of Everything Generation ability and largely enhance the generalization, we cannot get the class-priors of objects previously. Instead, the text token of `green' is an available and stable prompt to derive background without class-priors. For a `green' ($j$-th) text token, the corresponding weight is $\mathcal{A}_{j} \in \mathbb{R}^{H \times W}$. 
Our proposed green-background control mechanism is :
\vspace{-2mm}
\begin{equation}
\vspace{-1mm}
\mathcal{L}_{g} = \dfrac{1}{u}\sum_{l=1}^u|\mathcal{A}^l_{j}-(1-M)|_{mean},
\vspace{-1mm}
\end{equation}
where $M$ is GT mask of main object normalized to [0,1]. 
$\mathcal{A}^l_{j}$ represents cross-attention map corresponding to the $j$-th at the $l$-th cross-attention layer.
$u$ is total cross-attention layer.
We supervise cross-attention map $A_j$ to be close to the background segmentation mask $(1-M)$.
$mean$ is the pixel-level averaging.
\vspace{-3mm}
\subsection{Detailed-enhancement of Transition Boundary}
\vspace{-1mm}
To keep the detailed-object generation and avoid the crash of the transition boundary, we extract the high-frequency information (\textit{e.g.,} edge ) \cite{chen2023anydoor} to enhance the details generation around the transition boundary. 
\vspace{-2mm}
\begin{equation}
\vspace{-1mm}
H=(I\otimes S_x+I\otimes S_y)\odot I \odot M,
\vspace{-1mm}
\end{equation}
where $S_x$, $S_y$ means the horizontal and vertical Sobel kernels operation to filter the high-frequency priors. $I$ and $M$ are the input gray image and mask, respectively.  $\otimes$ refer to convolution operation and $\odot$ is the Haramard product on the pixel-level value. To guide the synthesis image to mimic the detailed objects, we restrict the synthesis image to high-frequency features closer to the GT images. 
\begin{equation}
\vspace{-1mm}
\mathcal{L}_{detail}=H_{\hat{z_0}}-H_{{z_0}},
\vspace{-1mm}
\end{equation}
where ${z_0}$ is the latent representation of an image $x_{0}$, and $\hat{z_0}$ is the estimated latent representation directly from $t$-th noising representation $z_t$ and the predicted noise $\epsilon_{\theta}(z_t,t,C)$ as follows:
\vspace{-2mm}
\begin{equation}
\vspace{-1mm}
\hat{z_0}=\frac{z_t-\sqrt{1-\bar{\alpha_t}}\epsilon_{\theta}}{\sqrt{\bar{\alpha}_{t}}}, \bar{\alpha}_{t} = \prod_{s=1}^{t}\alpha_s,
\vspace{-1mm}
\end{equation}
where $t$ means the $t$-th noise injection step, and $\bar{\alpha_t}$ is derived by predefined sequence of coefficients (${\alpha_t}$), controlling the variance schedule.

% \begin{equation}
% \bar{\alpha}_{t} = \prod_{s=1}^{t}\alpha_s.
% \end{equation}
\vspace{-2mm}
\subsection{Mask Generation and Refinement}
\vspace{-2mm}
% After generating t, 
We aim to get the corresponding matting-level annotations as well as the stable and pure green-screen-based objects. To this end, we firstly get the coarse mask $\tilde{M}_{cp}$ in the latent representation space via adding a matting header. 
% when the noise injection step $t<\delta T$ and $\delta=0.25$. 
The higher dimension (128) of the latent representation feature ($f_{\mathbf{decoder}}$)  is used to get the one-channel matting map in the VAE decoder as follows:
\vspace{-1mm}
\begin{equation}
\vspace{-1mm}
\tilde{M}_{cp}= \mathbf{ConR}(f_{\mathbf{decoder}}),
% \vspace{-1mm}
\end{equation}
where $\mathbf{ConR}$ is the matting-header with two stack of 2d convolution layer and SiLU activation function. Afterwards, we tie $\tilde{M}_{cp}$ with the GT ($M$) based on the dice and $L$1 loss as follows:
\vspace{-2mm}
\begin{equation}
\vspace{-1mm}
\mathcal{L}_{latent}=\frac{1}{N}\sum_{j=1}^N |\tilde{M}_{cp_j}-M_{j}|+1-\frac{2|\tilde{M}_{cp}\cap M|}{|\tilde{M}_{cp}|+|M|},
\vspace{-1mm}
\end{equation}
where N is total pixel amount, and $\cap$ is intersection of $\tilde{M}_{cp}$ and $M$. $||$ means pixel-level sum of a matrix. 
% After getting the $\tilde{M}_{cp}$, we use the K-Means clustering to get the dominant color of background and then fill the foreground zone with this dominant color. Such an accurate background improves the matting performance of following $\mathbf{GreenPost}$. 
Lastly we adopts a post-processing $\rm {GreenPost}$, using green-background priors to refine the pixel-level mask $\tilde{M}_{cp}$. The  $\rm {GreenPost}$ introduces the BackgroundMattingV2 \cite{lin2021real} as the post-processing after using the K-Means to cluster the background color calculated by the pixel-level mask $\tilde{M}_{cp}$ to fill the foreground pixel values. 
Then original image and accurate background are imported into the $\rm {GreenPost}$ to obtain the matting-level annotation $\tilde{M}_{p}$. 
\vspace{-1mm}
\begin{equation}
\vspace{-1mm}
\tilde{M}_{p}= {\rm GreenPost}(\tilde{M}_{cp}),
\vspace{-1mm}
\end{equation}

% adopts the BackgroundMattingV2 post-processing 
\vspace{-2mm}
\subsection{Objective Function}
\vspace{-2mm}
To control the pure and clean green-background, $\mathcal{L}_{g}$ is proposed to optimize the cross-attention feature for accurate fore- and background learning via a bidirectional manner. This setting is beneficial for the downstream $\rm{GreenPost}$ operation to get matting level annotations. 
In addition, to encourage the object with more details, a detailed-enhancement loss $\mathcal{L}_{detail}$ of transition boundary is used to facilitate the boundary details via high-frequency information (\textit{e.g.,} edge) alignment. 
The $\tilde{M}_{cp}$ is derived by the VAE decoder matting head, and $\mathcal{L}_{latent}$ are applied to mind the gap with GT denoted as $M$.

\begin{equation}
    \mathcal{L}_{total} = \mathcal{L}_{noise}+ \mathcal{L}_{g} + \mathcal{L}_{detail} + \mathcal{L}_{latent}.
\end{equation}

\section{Experiments}
\vspace{-2mm}
\subsection{Experimental Setups}
% \noindent\textbf{Datasets Description.} 
% Our Green100k dataset is collected and partially annotated by Adobe Premiere Pro Chromakey, where a part of hard cases with some shadow is carefully labeled by Adobe Photoshop. For more knowledge on the dataset refer to Section \secref{section:DataCollectionparts}. 
% \textcolor{red}{here}
% To verify that our model can be beneficial for the down-streaming matting task, we generate 10,000 synthesis portrait foreground images for the hair-level matting task, which suffers from privacy protection and labor-consuming annotation. Besides, following previous setting Adobe \cite{xu2017deep}, MS COCO \cite{lin2014microsoft} and Pascal VOC \cite{everingham2010pascal} are introduced to provide the background images for portrait image composition with synthesized foreground images. Besides, as the largest portrait matting set, P3M test set is used for further quantitative analysis. 

\noindent\textbf{Training Details.} Stable Diffusion V1.5 is adopted as the pretrained model. 
% Our model undertakes two tasks including the object generation on the pure green background which facilitates the following matting-level annotation tasks. 
Our model undertakes two tasks including the object generation on the pure green background and matting-level annotation tasks.
First, we only train a DiffuMatting generation model without matting parameters to ensure the pure and stable green-screen generation with a learning rate of 2e-6 and batch size of 2 on the two NVIDIA V100 GPUs for 1.5 days. Then, we use the well-trained parameters in the last step and jointly train the generation and matting stage on the two NVIDIA V100 GPUs for 1 day. This setting can avoid the misalignment of green-screen generation at the beginning of training, which leads to the deterioration of matting parameters training. 
% #
\begin{table}[t]
\scriptsize
\centering
\caption{Green-screen generation quality evaluation.}\label{tab:tab1_gsg}
\vspace{-3mm}
\setlength\tabcolsep{10.5pt}
\begin{tabular}{r|c|c|c}
%\begin{tabular}{ccccc}
\toprule
% \multicolumn{3}{c|}{Configurations}  & Performance  \\ \hline 
 Metric & LoRA & Dreambooth  & Ours \\  \hline
$\mathbf{GSG}\downarrow$  & 134.17& 133.34 & \textbf{98.98}  \\ 
$A_s\uparrow$   & 4.78 & 4.85 & \textbf{5.26} \\    
% $M\downarrow$   & 0.029 &  0.025 & \textbf{0.024} \\    
% $E_\phi\uparrow$   & 0.898  & 0.926 & \textbf{0.932} \\    %\hline
\bottomrule
\end{tabular}
\vspace{-3mm}
\end{table}

\begin{figure}[t!]
\centering
\includegraphics[width=0.95\linewidth]{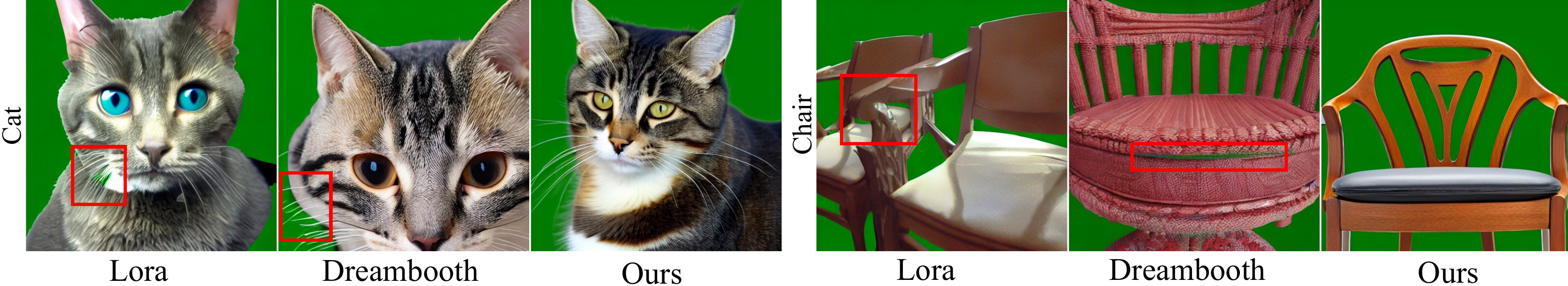}
\vspace{-3mm}
\caption{
Visual performance of our DiffuMatting on green-screen-based object generation in comparison with LoRA and Dreambooth fine-tuning in our Green100K.
}
% \vspace{-4mm}
\label{fig:basline_lorad}
\end{figure}

\begin{figure*}[t!]
\centering
\includegraphics[width=0.9\textwidth]{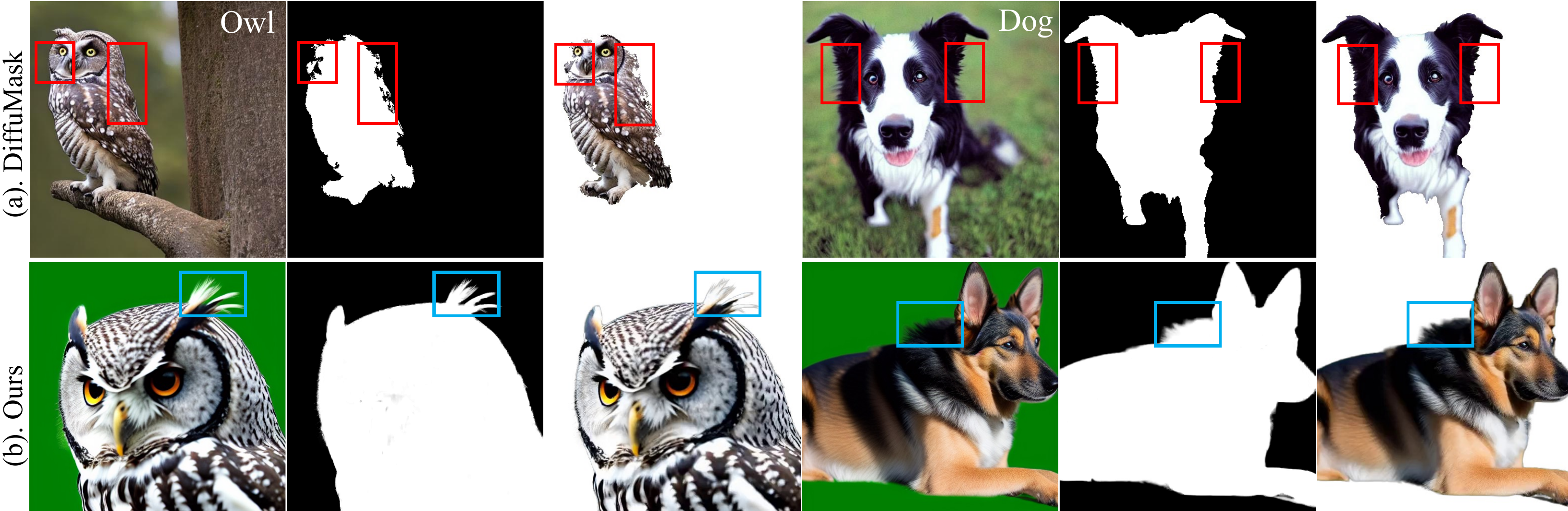}
\vspace{-2mm}
\caption{Matting-level annotation analysis. Our matting-level annotation (blue box) \textit{vs.} pixel-level mask generated by DiffuMask \cite{wu2023diffumask} (red box) in the class of dog and Owl (bird). 
Each class-object generation by DiffuMask requires fine-tuning AffineityNet for this specific class to get post-processing annotation results.
}\label{fig:fig_matting_w_diffumask}
% \vspace{-5mm}
\end{figure*}

\noindent\textbf{Evaluation Metric.}
% To evaluate the pure green-screen generation, we propose a K-means clustering to estimate the main dominant color. 
Considering no existing available criteria to evaluate the quality of stable and pure green-screen background, we propose a green-screen generation quality (GSG) that uses the K-Means to cluster the dominant color of synthesis images and calculate the distance with the green-screen color.
\vspace{-2mm}
\begin{equation}
\vspace{-2mm}
\mathbf{GSG}=|({\mathop{\arg\min}\limits_{S}\sum_{i=1}^k\sum_{x\in S_i} ||\mathbf{x}-\mathbf{\mu}_{i}||^{2}})_{\mathrm{top}}-P_g|_{Eu},
% \vspace{-1mm}
\end{equation}
where x is a set of observations divided into k sets. $\mu_{i}$ is the mean of points in $S_{i}$. $()_{\mathrm{top}}$ denotes the top set $S$, the dominant color of an image. $P_g$ is the GT geometric coordinates of pure green-screen color and $||_{Eu}$ is the Euclidean distance between the estimated dominant color and GT coordinates.
Following previous works, the sum of absolute differences (SAD), mean squared error (MSE),
% mean absolute difference (MAD), 
gradient (Grad.), and Connectivity (Conn.) are used.

% \begin{figure}[t!]
% \centering
% \includegraphics[width=0.85\linewidth]{figures/crop_green_screen_quality_v2.pdf}
% \vspace{-3mm}
% \caption{
% Visual performance of our DiffuMatting on green-screen-based object generation in comparison with Lora and Dreambooth fine-tuning in our Green100K.
% }
% \vspace{-2mm}
% \label{fig:basline_lorad}
% \end{figure}
\vspace{-2mm}
\subsection{Consistency Green-Screen Generation Capability}
\vspace{-2mm}
To verify the stability, consistency, and pure green-screen generation, we compare our proposed method with LoRA and Dreambooth models fine-tuning in the same training set Green100K with the default setting.
% including learning rate, \textit{etc.} 
Specifically, we randomly generate 10 class images (\textit{e.g.,} dog, cat, woman, man, bike, chair, bridge, laptop, desk, car) and each class contains 10 images. The aesthetic-score $A_s$ \cite{schuhmann2022laion} is to analyze the aesthetic score of the green-screen images. 
The quantitative results are listed in Table \ref{tab:tab1_gsg}. Our training setting powered by the green-background control is much better than LoRA and Dreambooth models. 
Some qualitative results are also shown in Fig. \ref{fig:basline_lorad}. DiffuMatting shows a strong generation ability on the details of transition area (cat whiskers) and the purity of green-screen. 

% $$

%%%%-------------
% \begin{table}[t!]
% \footnotesize
% \centering
% \caption{Green-screen generation quality evaluation.}\label{tab:tab1_gsg}
% % \vspace{-10pt}
% % \renewcommand{\arraystretch}{0.5}
% \setlength\tabcolsep{10.5pt}
% \begin{tabular}{r|c}
% %\begin{tabular}{ccccc}
% \toprule
% % \multicolumn{3}{c|}{Configurations}  & Performance  \\ \hline 
%  Metric & Lora  \\  \hline
% $\mathbf{GSG}\downarrow$  & 134.17  \\ 
% $A_s\uparrow$   & 4.78  \\    
% % $M\downarrow$   & 0.029 &  0.025 & \textbf{0.024} \\    
% % $E_\phi\uparrow$   & 0.898  & 0.926 & \textbf{0.932} \\    %\hline
% \bottomrule
% \end{tabular}
% % \vspace{-5mm}
% \end{table}
%%%%--------
\vspace{-3mm}
\subsection{Matting-level Annotation Analysis}

% Art design. Our DiffuMatting is acted as the base-model, and a style-based Lora can be trained by the style training dataset. Thus, this combination can provide the user-specified style 4-channel images with high-accurate alpha, as shown in Fig. \ref{fig:qinghuaci}.

\begin{figure*}[t!]
\centering
\includegraphics[width=0.9\textwidth]{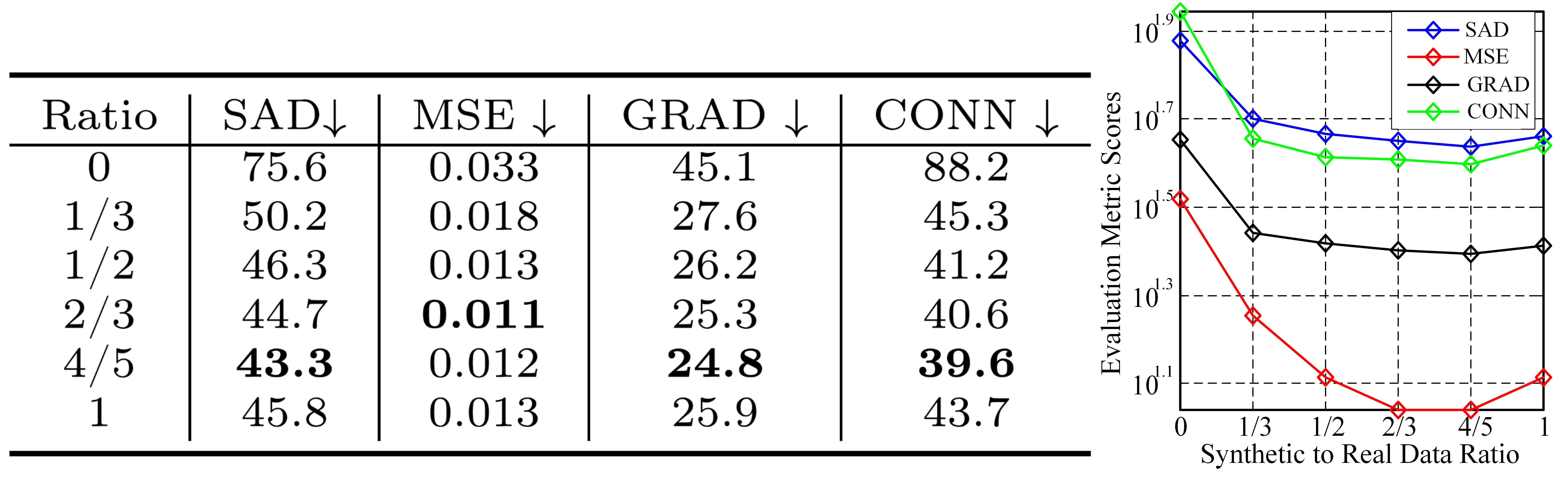}
\vspace{-3mm}
\caption{The general matting results using our synthesis 10K General-matting set based on the Indexnet \cite{lu2019indices}.
}\label{fig:general_matting}
\vspace{-3mm}
\end{figure*}

% \begin{figure*}[t!]
% \centering
% \includegraphics[width=0.9\textwidth]{figures/229_portrait.png}
% \vspace{-3mm}
% \caption{The portrait matting results using our synthesis 10K Portrait-matting set based on the P3M-Net \cite{li2021privacy}.
% }\label{fig:portrait_matting}
% \vspace{-5mm}
% \end{figure*}

As a recent-related work \cite{wu2023diffumask}, DiffuMask is a diffusion model to synthesize pixel-level annotations. 
% Different from DiffuMask, our DiffuMatting can achieve \textit{1).} Matting-level annotation accuracy. \textit{2).} Anything generation not restricted to the class-priors in the training set. 
DiffuMask fails to provide a matting-level annotation and suffers from the misalignment of synthesized images and annotations to a certain extent. Each class-object generation requires fine-tuning AffineityNet for this specific class to get post-processing annotation. DiffuMask released two class AffineityNet weights (\textit{i.e.}, dog and bird). We compare matting-level annotations of DiffuMatting with the pixel-level masks in Fig. \ref{fig:fig_matting_w_diffumask}. It shows that our DiffuMatting achieves hair-level annotation (\textit{e.g.,} fur, hair) compared with DiffuMask.

\begin{figure*}[t!]
\centering
\includegraphics[width=0.9\textwidth]{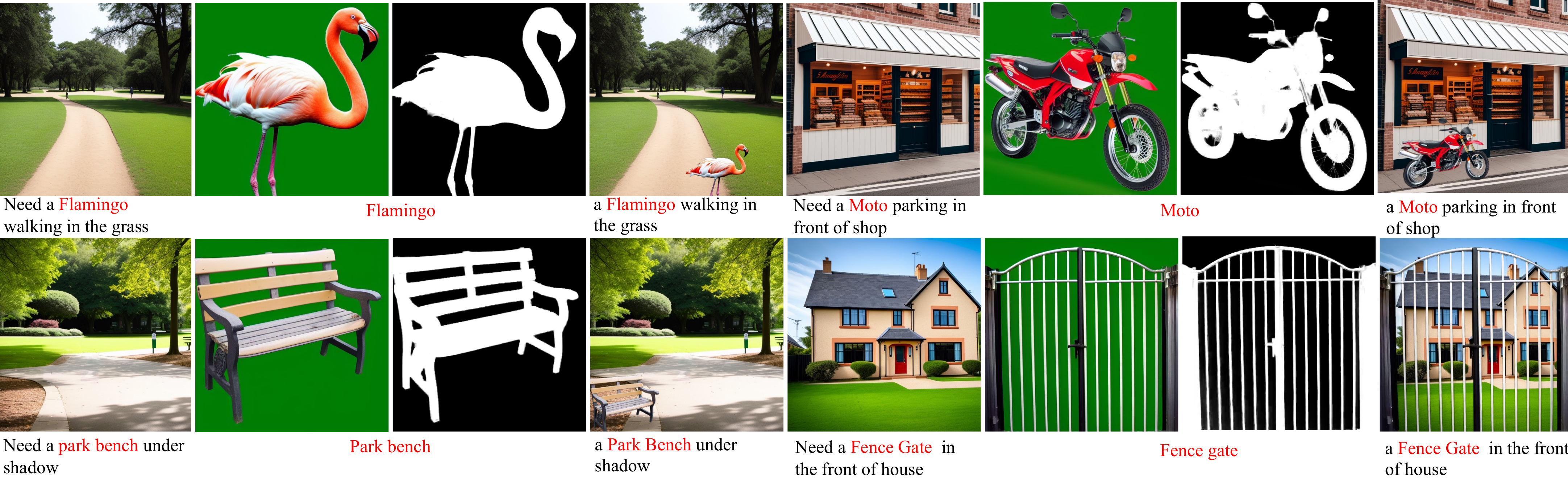}
% \vspace{-3mm}
\caption{Image Composition. DiffuMatting generates green-screen objects, copies the object with the matting-level annotation, and then pastes it into desired scenario.
}\label{fig:composition}
% \vspace{-3mm}
\end{figure*}

\begin{figure*}[t!]
\centering
\includegraphics[width=0.9\textwidth]{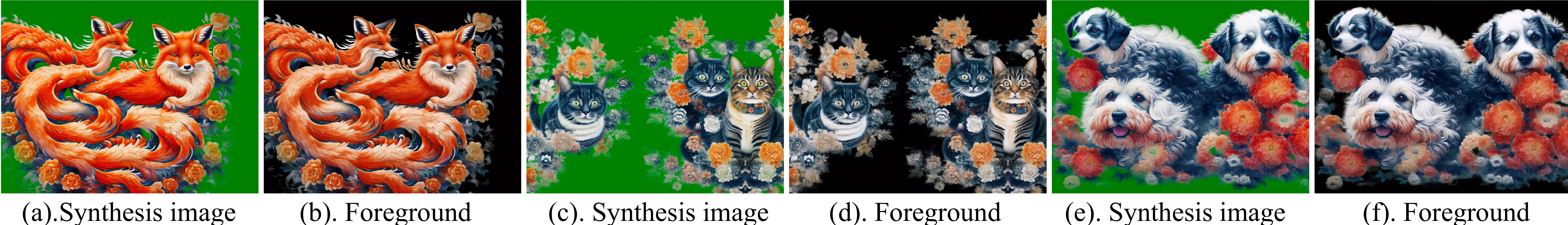}
% \vspace{-1mm}
\caption{Art design: 
Style of blue and white porcelain texture with alpha-channel generated by DiffuMatting.
}\label{fig:art_design}
% \vspace{-5mm}
\end{figure*}

% \vspace{-2mm}
\subsection{Applications}
% \vspace{-2mm}
\noindent\textbf{Downstream Matting Task.}
To verify the benefits of our matting-level annotation to the down-streaming matting-level task, we analyze our synthesized images 
% on two mainstream matting tasks (General-object and Portrait matting). 
on General-object matting. 
\noindent \textit{1).} General Object Matting: We synthesize the 10K general-object matting set (GOM) based on the around 500 object classes guided by the detailed captions of Green100K dataset. We train a general matting model based on the commonly-used Indexnet \cite{lu2019indices} under the same default setting on various training sets, consisting of images from the general-object matting set (GOM) and Deep Image Matting (DIM) dataset. The synthetic to real data ratio $\gamma$ denotes that $\gamma$ ratio of images from the DIM dataset and ($1- \gamma$) ratio of images from the synthesized GOM. As shown in Fig. \ref{fig:general_matting}, the combined dataset (highlighted in bold) used for matting gets better results (15.4\% relative error lower on MSE, 9.4\% on CONN) than only using the DIM dataset (Ratio $\gamma$=1), indicating that synthesized images by our DiffuMatting is useful for downstream general object matting even though it stills exists domain difference between the real and synthesized dataset. 
\begin{figure*}[t!]
\centering
\includegraphics[width=0.9\textwidth]{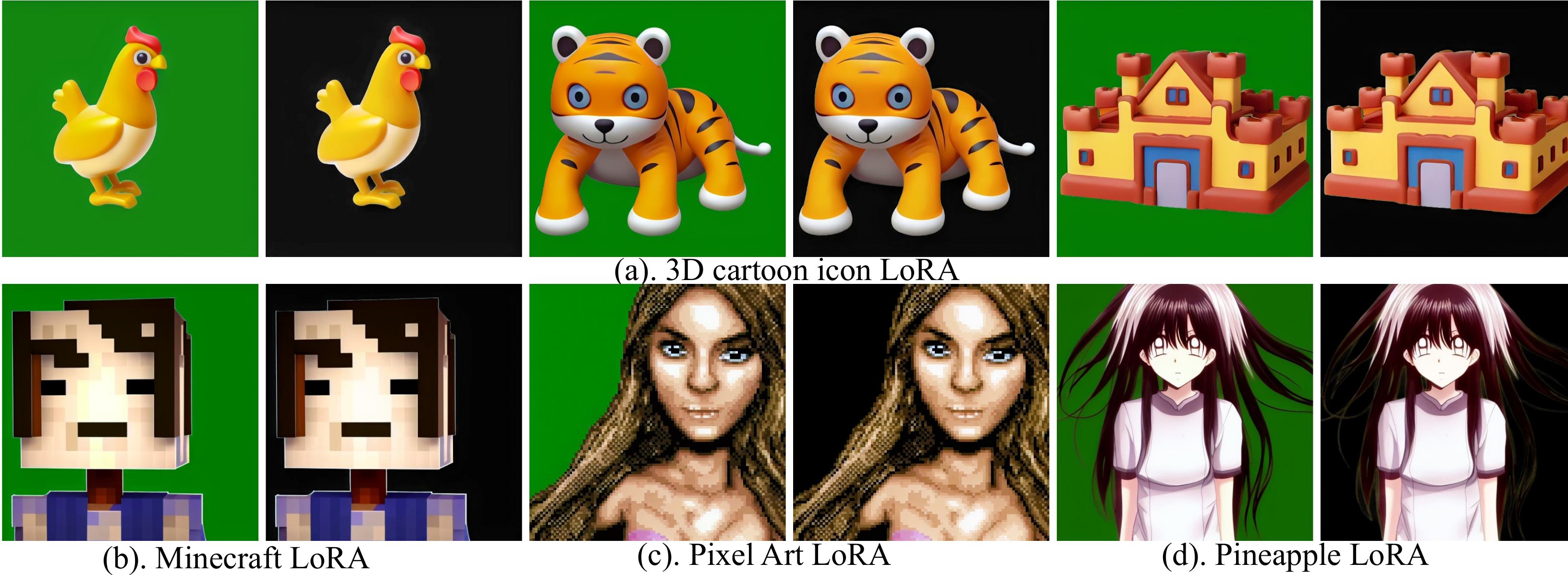}
% \vspace{-3mm}
\caption{Applying to Community LoRAs: we select some typical style LoRAs (\textit{i.e.,} 3D cartoon icon LoRA, Minecraft LoRA, Pixel Art LoRA, Pineapple LoRA) for testing. 
% It shows that the style quality of generated images is kept, and  the alpha channel can be provided to extract the foreground object.
}\label{fig:art_design_lora_exp}
% \vspace{-3mm}
\end{figure*}

\begin{figure*}[t!]
\centering
\includegraphics[width=0.9\textwidth]{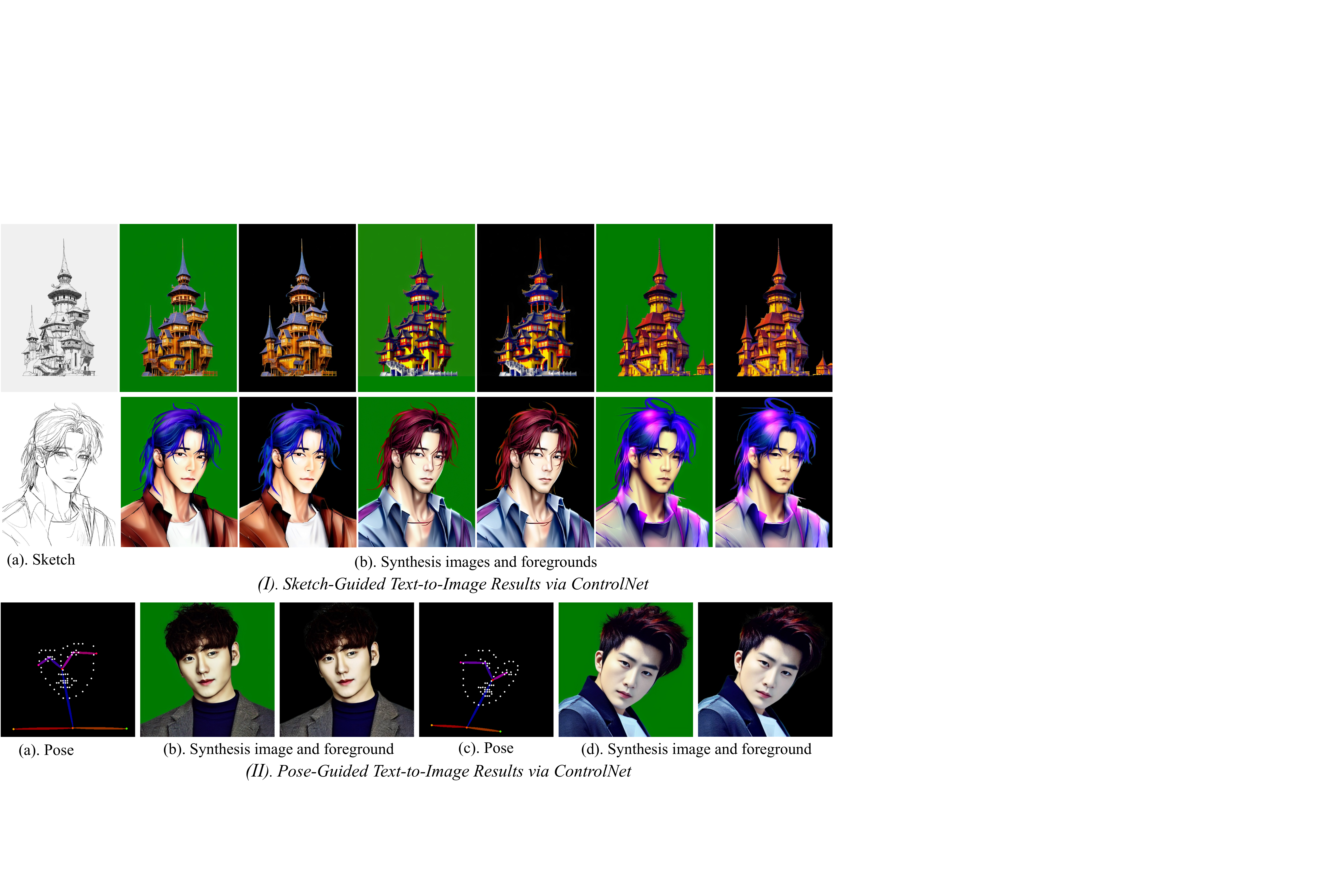}
% \vspace{-2mm}
\caption{Controllable text-to-image generation combing with the ControlNet. Our Diffumatting can be directly combined with the ControlNet for sketch-guided and pose-guided text-to-image generation. 
% Style of blue and white porcelain texture in 4-channels with alpha-mask generated by our DiffuMatting.
}\label{fig:art_design_control}
% \vspace{-5mm}
\end{figure*}

% \noindent \textit{2).} Portrait Matting: We also generate the 10K human-portrait-matting data (PM) including various age stages, races, genders, and nationalities.  The largest public portrait dataset available is the P3M dataset with 10K images.

% This kind of portrait data is usually strictly restricted by privacy-preserving standards and labor-consuming costs. Thus most cases of the P3M dataset usually blur the face part to avoid the privacy conflict. The amount of sharp faces is only 500, and the rest are blurred faces. 
% Our synthesized image naturally avoids the leakage of the user's identity privacy. 

% The quantitative results are shown in Fig. \ref{fig:portrait_matting}. 
% Adding our synthesis 10K portrait-matting set consistently improves the performance of network, and the optimal ratio is $4/5$ meaning that $1/5$ from the synthesized PM set and $4/5$ from the real P3M dataset. The ratio setting decreases the MSE error by 11.4\% than the results of only using the real set. 

\noindent\textbf{Image Composition: Generate, Copy and Paste.} Image composition usually suffers from the coarse mask that does not describe the edges and holes of objects. The coarse mask can be generated by the segmentation algorithms (\textit{e.g.,} SAM) in a low-cost manner. In contrast, matting-level annotations require much larger labor-consuming cost which is not cheaply available for image composition. But our DiffuMatting greatly decreases the cost of matting annotations benefiting from its great generalization that supports generating `anything' matting-level annotations. Thus, our DiffuMatting can generate green-screen objects, copy the object with the matting-level annotation, and paste the object into composition scenarios in a closed loop. Some image composition results can be seen in the Fig. \ref{fig:composition}, adding different elements to a target scene.

\begin{table}[t!]
\scriptsize
\centering
\caption{Ablation study. The quality evaluation of green-screen synthesis. GB is denoted as the green-background control loss, and DE is detailed-enhancement loss.}\label{tab:tab1_abl}
\vspace{-5pt}
\setlength\tabcolsep{10.5pt}
\begin{tabular}{r|c|c|c}
%\begin{tabular}{ccccc}
\toprule
% \multicolumn{3}{c|}{Configurations}  & Performance  \\ \hline 
 Metric & \textit{w/o} GB & \textit{w/o} DE & Ours \\  \hline
$\mathbf{GSG}\downarrow$  & 119.8&  112.5 & \textbf{98.98}  \\ 
$A_s\uparrow$   & 5.02 &  4.95 & \textbf{5.26} \\    
% $M\downarrow$   & 0.029 &  0.025 & \textbf{0.024} \\    
% $E_\phi\uparrow$   & 0.898  & 0.926 & \textbf{0.932} \\    %\hline
\bottomrule
\end{tabular}
\vspace{-2mm}
\end{table}

\begin{figure}[t!]
\centering
\includegraphics[width=0.9\linewidth]{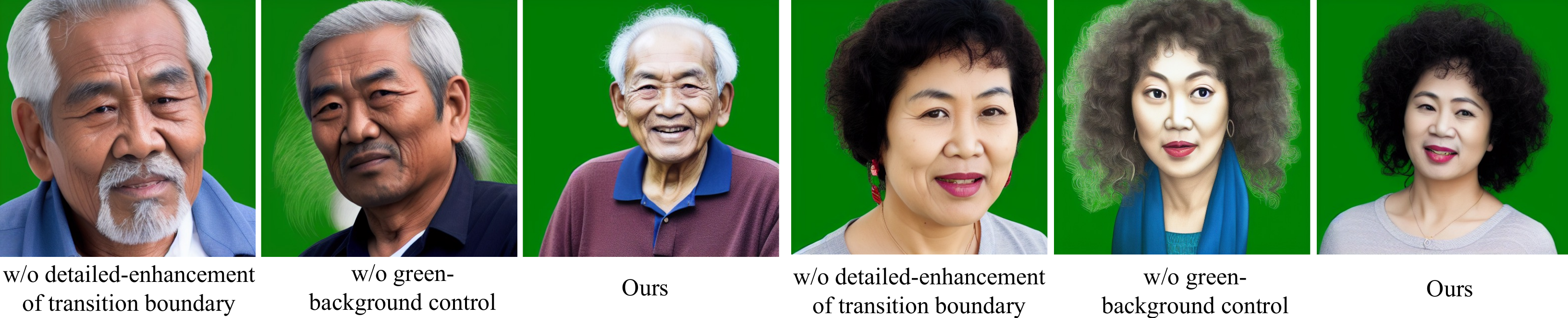}
% \vspace{-3mm}
\caption{
Visual quality of ablation study with different configurations. 
}
\label{fig:ablationstudy}
% \vspace{-3mm}
\end{figure}

\begin{figure}[t!]
\centering
\includegraphics[width=0.9\linewidth]{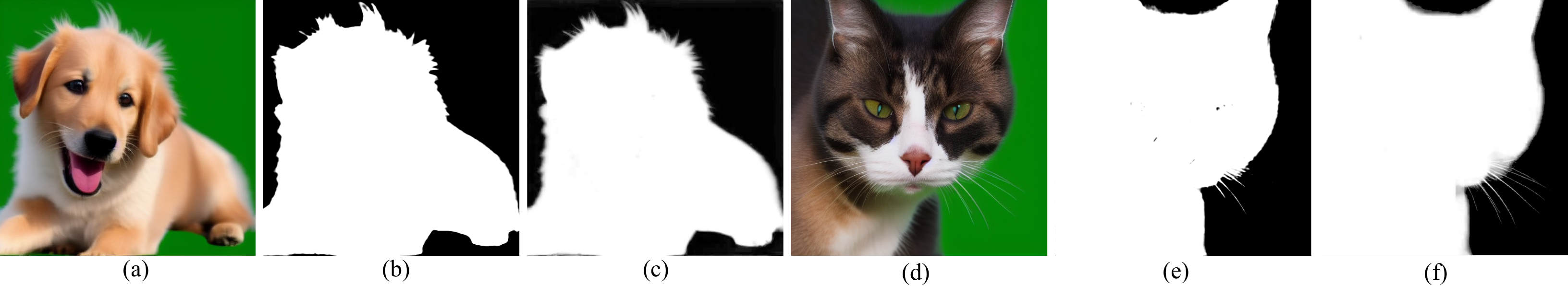}
% \vspace{-3.5mm}
\caption{
The matting refinement refines the pixel-level mask to the matting-level annotation. (a) and (d) green-screen synthesized images, (b) and (e) pixel-level masks from the matting-head of latent space, (c) and (f) the outputs of the ${\rm GreenPost}$ matting refinement process.
}
\label{fig:mattingrefine}
% \vspace{-5mm}
\end{figure}

\noindent\textbf{Community-friendly Art Design and Controllable Generation.} DiffuMatting is well compatible with LoRA models, and it provides the potential for users to customize the specific-style RGBA 4-channels images. 
DiffuMatting acts as the base model and a style-based LoRA can be trained by the users' style images. 
% Thus, this combination can provide the user-specified 4-channel style images with high-accurate alpha. 
The self-training style-LoRA results of blue and white porcelain texture in 4-channels are shown in Fig. \ref{fig:art_design}. Besides, our DiffuMatting can be applied to the various community LoRAs models without any additional training process. For example, Minecraft LoRA, 3D cartoon icon LoRA, pixel art LoRA, and Pineapple LoRA are used to get diverse results in Fig. \ref{fig:art_design_lora_exp}. 
We see that applying the existing Community LoRA can still keep the style quality, and get a good transparency channel to extract the foreground objects. 
The existing control models (\textit{e.g.,} ControlNet) can be directly applied in DiffuMatting to enrich the controllable image editing functionality in Fig. \ref{fig:art_design_control}. For sketch-guided text-to-image generation, our model can generate editable style images (\textit{e.g.,} Western and Asian style castle, or the specified color of hair) according to the ControlNet signal. Given human pose knowledge, our model can also edit the portrait gesture guided by the ControlNet. 
%
%
%
% \noindent\textbf{Community}

% \noindent\textbf{3D Reconstruction.}
% Our accurate annotation can also provide the detailed topological structure which facilitates the following 3D geometric modeling.

% \vspace{-2mm}
\subsection{Ablation Study}
% \vspace{-2mm}
\noindent\textbf{Green-background Control.} 
Fig. \ref{fig:ablationstudy} shows that the setting without green background control leads to the color bleeding around the transition area of fore- and background, \textit{e.g.,} the white hair color of elder affecting the green-screen background and the transition hair area of woman is mixed by the green and hair color. Also in Table \ref{tab:tab1_abl}, the setting without green background control gets a much worse GSG score, which indicates that the green-screen background is not satisfactory. 
It also deteriorates the aesthetic-score in such a setting.

\noindent\textbf{Detailed-enhancement of transition boundary.} 
As shown in Fig. \ref{fig:ablationstudy}, the setting without the detailed-enhancement of transition boundary suffers from the detail loss of the boundary, 
% between the fore- and background, 
especially on the hair area. After adding both green-background control and detailed-enhancement of transition boundary, the detailed-preserving of fur or hair and distinction of background and foreground are
% largely 
improved to steadily generate high-quality synthesis image and matting annotation. From Table \ref{tab:tab1_abl}, we find that the detailed-enhancement loss has a greater impact on the aesthetic score compared to green-background loss.

\noindent\textbf{Mask Generation and Refinement.} Our matting head can acquire a pixels-level mask which does not satisfy our matting-level annotation. Then the matting refinement process guided by the background-prior is proposed to refine the pixels-level mask. 
As shown in Fig. \ref{fig:mattingrefine},  after the $\rm {GreenPost}$ matting refinement process, the matting output shows more transparent information.
% \textit{e.g.,} the fur of a dog or cat shows much better edge texture.

\begin{figure}[t!]
\centering
\includegraphics[width=0.9\linewidth]{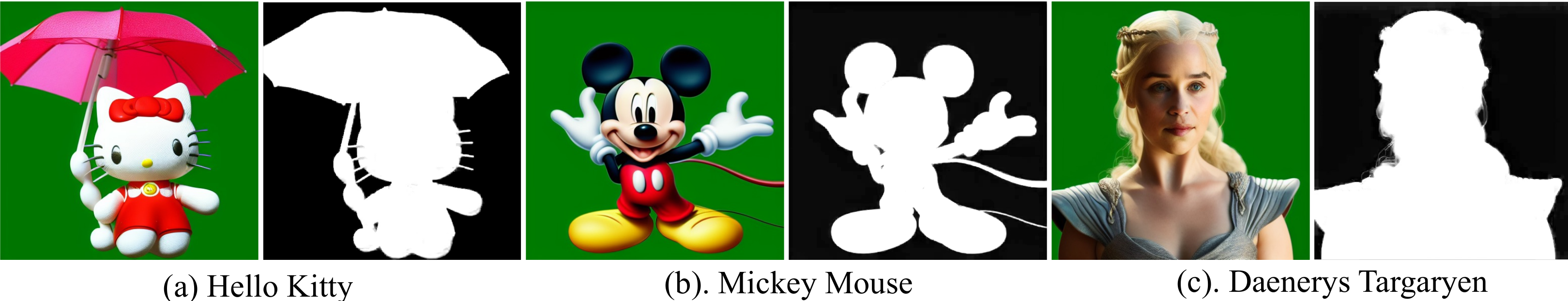}
\vspace{-3mm}
\caption{
Anything Generation Ability. The object generation (\textit{e.g.,} cartoon and celebrity) beyond the Green100K.
}
\label{fig:anything}
\vspace{-5mm}
\end{figure}

\noindent\textbf{Strong Generalization beyond Green100K.} 
Green100K is a teacher dataset that teaches the diffusion model to paint on a fixed green-screen canvas. After learning this principle, our model can also inherit the Anything Generation Ability of pretrained large-scale diffusion model. 
As shown in Fig. \ref{fig:anything}, the images and matting-level annotations largely beyond Green100K (\textit{e.g.,} Hello Kitty, Mickey Mouse, or even Daenerys Targaryen) can be easily and steadily generated. 

% \vspace{-2mm}
\section{Conclusion}
% \vspace{-2mm}
We propose a DiffuMatting taught by our Green100k to paint on a green-screen canvas, which ingeniously separates the foreground.
With the aid of the background, transition boundary control, and matting refinement, DiffuMatting combines the robust generative capabilities of diffusion and the added functionality of `matting anything'. 
As such, our DiffuMatting serves as a highly accurate `anything matting' factory which facilitates the down-streaming matting task, while also achieving the composition and  creation of content or community-friendly art design and controllable generation.

% bridging the gap to generate, replicate, and integrate ultra-detailed objects and their corresponding matting annotations into any scene.
% Image Composition,Community-friendly Art Design and Controllable Generation

% \noindent \textbf{Limitation and Social Impacts.}

\noindent\textbf{Limitation and Social Impacts.} Our work is primarily centered around the generation on the matting-level foreground of `Anything'. It has limitations with simultaneously synthesizing non-green screen images and matting-level annotations without any assistance of image inpainting technique. DiffuMatting technology can be used for content creation and carry risks in illicit industries. To mitigate potential misuse, explicit markings will be applied to generated images. 

\noindent\textbf{Acknowledgments.} 
This work was supported by National Key R\&D Program of China (No. 2022ZD0118202), in part by the National Natural Science Foundation of China (No. 62072386), in part by Yunnan Provincial Major Science and Technology Special Plan Project (No. 202402AD080001),  in part by by Henan Province key research and development project (No. 231111212000) and the Open Foundation of Henan Key Laboratory of General Aviation Technology (No. ZHKF-230212).

\clearpage  % TODO REVIEW/FINAL: This \clearpage needs to be removed from both review and camera-ready versions.

% ---- Bibliography ----
%
% BibTeX users should specify bibliography style 'splncs04'.
% References will then be sorted and formatted in the correct style.
%
\bibliographystyle{splncs04}
\bibliography{main}
\end{document}